\documentclass[letterpaper]{article} 
\usepackage{aaai2026}  
\usepackage{times}  
\usepackage{helvet}  
\usepackage{courier}  
\usepackage[hyphens]{url}  
\usepackage{graphicx} 
\usepackage{natbib}  
\usepackage{caption} 
\usepackage{algorithm}
\usepackage{algorithmic}

\usepackage{newfloat}
\usepackage{listings}
\DeclareCaptionStyle{ruled}{labelfont=normalfont,labelsep=colon,strut=off} 
\floatstyle{ruled}
\newfloat{listing}{tb}{lst}{}
\floatname{listing}{Listing}
\title{GraphIC: A Graph-Based In-Context Example Retrieval Model\\ for Multi-Step Reasoning}
\author{
    Jiale Fu\textsuperscript{\rm 1,2},
    Yaqing Wang\textsuperscript{\rm 3}\footnotemark[1],
    Simeng Han\textsuperscript{\rm 4},
    Jiaming Fan\textsuperscript{\rm 1,2},
    Xu Yang\textsuperscript{\rm 1,2}\footnote{Corresponding author}\\
}
\affiliations{
    \textsuperscript{\rm 1}Southeast University\\
    \textsuperscript{\rm 2}Key Laboratory of New Generation Artificial Intelligence Technology and Its Interdisciplinary Applications (Southeast University), Ministry of Education, China.\\
    \textsuperscript{\rm 3}Beijing Institute of Mathematical Sciences and Applications\\
    \textsuperscript{\rm 4}Stanford University

    jiale.fu@seu.edu.cn, wangyaqing@bimsa.cn, shan6@stanford.edu, \{jiaming.fan, xuyang\_palm\}@seu.edu.cn
}

\usepackage{amssymb}
\usepackage{amsmath}
\usepackage{booktabs}
\usepackage{multirow}
\usepackage{makecell}
\usepackage{subcaption}
\usepackage{enumitem}

\newlist{mylist}{enumerate*}{1}
\setlist[mylist]{label=(\arabic*),itemjoin=\ }

\begin{document}

\maketitle

\begin{abstract}
In-context learning (ICL) enhances large language models (LLMs) by incorporating demonstration examples, yet its effectiveness heavily depends on the quality of selected examples. Current methods typically use text embeddings to measure semantic similarity, which often introduces bias in multi-step reasoning tasks. This occurs because text embeddings contain irrelevant semantic information and lack deeper reasoning structures. To address this, we propose \textbf{GraphIC}, a graph-based retrieval model that leverages reasoning-aware representation and specialized similarity metric for in-context example retrieval. GraphIC first constructs \textit{thought graphs}—directed, node-attributed graphs that explicitly model reasoning steps and their dependencies—for candidate examples and queries. This approach filters out superficial semantics while preserving essential reasoning processes. Next, GraphIC retrieves examples using a novel similarity metric tailored for these graphs, capturing sequential reasoning patterns and asymmetry between examples. Comprehensive evaluations across mathematical reasoning, code generation, and logical reasoning tasks demonstrate that GraphIC outperforms 10 baseline methods. Our results highlight the importance of reasoning-aware retrieval in ICL, offering a robust solution for enhancing LLM performance in multi-step reasoning scenarios.
\end{abstract}

\begin{links}
    \link{Code}{https://github.com/jialefu/GraphIC}
\end{links}

\section{Introduction}

In-context learning (ICL) \cite{NEURIPS2020_1457c0d6,peng2024live,jiang2025mimic} allows large language models (LLMs) to adapt to new tasks by incorporating a few demonstration examples within the input prompt, without updating model parameters. However, studies reveal that ICL performance heavily depends on the quality of selected in-context examples (ICEs) \cite{zhao2021calibrate, yang2023exploring,li2024configure}, motivating extensive research on ICE selection strategies. Current approaches \cite{liu2022makes, rubin2022learning} typically use text embeddings to measure semantic similarity between queries and candidate examples, achieving success in semantic-centric tasks like text classification and translation \cite{agrawal2023context}.

\begin{figure*}[t]
    \centering
    \includegraphics[width=.97\linewidth]{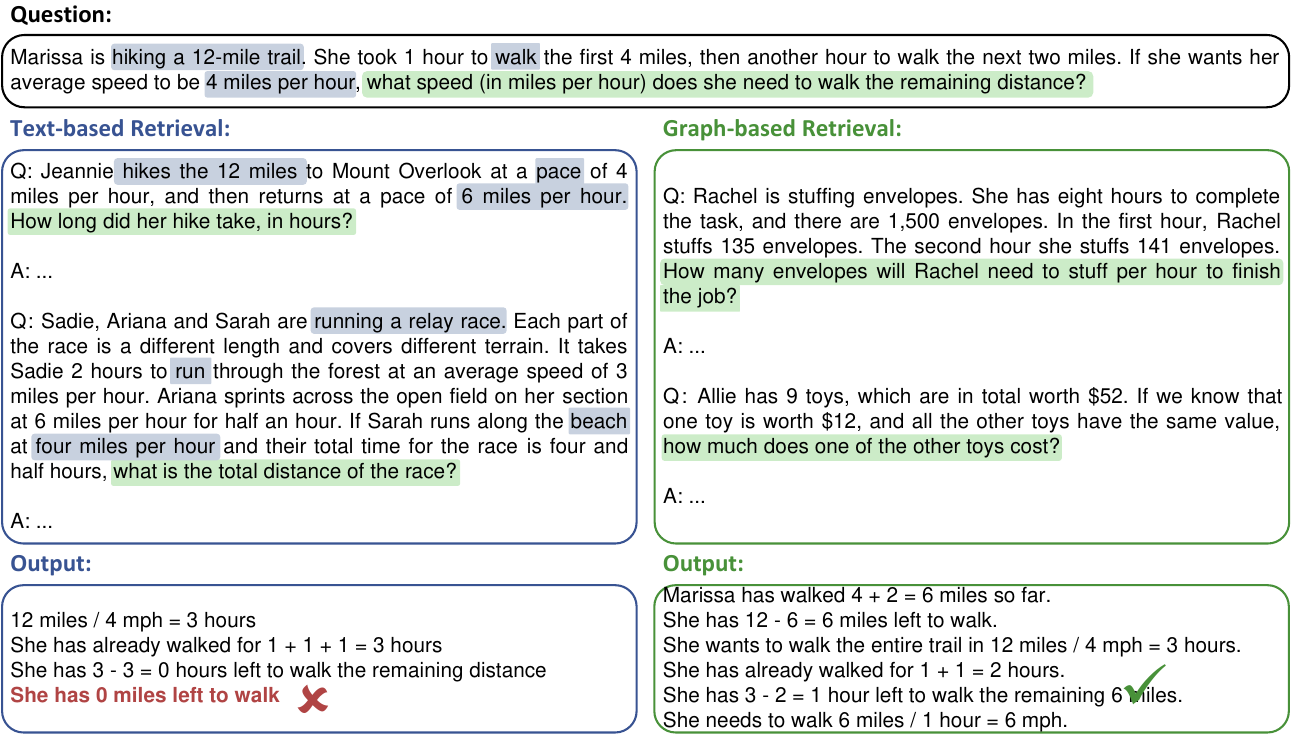}
    \caption{ICL with different ICE retrieval mechanisms. The left panel shows examples retrieved via BERT embedding \cite{devlin-etal-2019-bert}, while the right panel displays examples retrieved via GraphIC. Semantically related terms are highlighted in blue, and quantities or variables needing resolution are highlighted in green.}
    \label{fig:example}
\end{figure*}

However, these text-based methods face significant limitations in multi-step mathematical and logical reasoning tasks. This is because text embedding encodes a substantial amount of shallow semantic information, which is irrelevant to the underlying reasoning processes. This extraneous information introduces bias in the selection of ICEs \cite{an2023skill}. As shown in Figure~\ref{fig:example} (left), when solving a speed calculation problem, text-based methods may retrieve examples about distance/time calculations due to shallow semantic similarity, inducing incorrect reasoning paths (e.g., calculating distance instead of speed). This observation motivates our key insight: Effective ICE selection for reasoning tasks requires representations that explicitly model cognitive processes rather than textual surface forms.

Drawing from cognitive science \cite{friston2008hierarchical} and graph-based reasoning works \cite{besta2024graph, yao2024tree}, we propose \textit{thought graphs}—directed node-attributed graphs where nodes represent reasoning steps and edges denote step dependencies (i.e., a child step can only proceed after the parent step is completed). This representation filters irrelevant semantics while preserving essential reasoning patterns. An example of a thought graph is shown in Figure~\ref{fig:graph_frr}~(a).

Furthermore, we propose a similarity metric for thought graphs tailored to multi-step reasoning tasks. Specifically, we introduce a parameter \( W \) to model the reasoning pattern of a given process. Given two thought graphs, \( G_1 \) and \( G_2 \), we first estimate the reasoning pattern parameter \( W_1 \) for \( G_1 \), then assess its applicability to \( G_2 \). This applicability serves as the similarity measure \( s(G_1, G_2) \). Our approach naturally captures the sequential structure of reasoning steps and accounts for the inherent asymmetry between examples, which we discuss in detail in Section~\ref{sec:similarity}.

Building upon these foundations, we introduce GraphIC, a \textbf{Graph}-based \textbf{I}n-\textbf{C}ontext Example Retrieval Model. GraphIC achieves reasoning-aware example selection through three key phases: (1) constructing thought graphs for both the query and candidate examples, (2) calculating graph similarity using our proposed metric, and (3) retrieving the top-$k$ most relevant examples. As illustrated in Figure~\ref{fig:example} (right), GraphIC effectively identifies examples that align with the reasoning process (such as calculations for the number of envelopes to stuff per hour or unit price), even if they lack semantic similarity to the query (e.g., speed computation). This enables the LLM to solve the problem correctly.

Through comprehensive evaluations across mathematical reasoning, code generation, and logical reasoning tasks, GraphIC demonstrates superior performance over 10 baseline methods including both training-free and training-based approaches. To sum up, our key contributions are:

\begin{enumerate}
    \item \textbf{A representaion.} We introduce a novel graph-based representation, called \textit{thought graph}, to model multi-step reasoning processes. This representation effectively filters out irrelevant shallow semantic information while preserving the essential reasoning steps.
    \item \textbf{A similarity metric.} We introduce a similarity metric for thought graphs tailored to multi-step reasoning tasks that capture the sequential nature of the steps and the asymmetry between examples.
    \item \textbf{Empirical validation.} Our experimental results indicate that GraphIC, despite being a training-free model, outperforms both training-free and training-based models across various multi-step reasoning tasks.
\end{enumerate}

\section{Related Work}

Existing ICE selection techniques can be classified as either training-free or training-based, depending on whether a retriever needs to be trained.

Training-free approaches are generally divided into two types: (i) those that use heuristic criteria such as similarity~\cite{liu2022makes, hu2022context}, diversity~\cite{cho2023prompt, zhangautomatic, levy2023diverse, hongjin2022selective, zhang2023ideal}, complexity~\cite{fu2022complexity}, or combinations of these~\cite{agrawal2023context, tonglet2023seer, gupta2023coverage} to select in-context examples (ICEs); (ii) those that leverage feedback from LLMs, such as probability distributions~\cite{wu2023self, nguyen2023context, li2023finding, yang2023representative}, perplexity~\cite{gonen2023demystifying}, or the model's generated output~\cite{an2023skill} to guide the selection process. Training-free approaches eliminate the computational and time costs of model training, but their simpler architecture often leads to lower performance than training-based methods.

Training-based methods are generally divided into two types. The first learns to select individual examples and then extends this to $k$-shot scenarios~\cite{rubin2022learning, xiong2024dqlore, guptagistscore}. The second models the selection of a group of examples as a whole~\cite{ye2023compositional, wang2024large, zhang2022active, scarlatos2023reticl, lu2022dynamic, xureprompting, yang2024lever}. Training-based approaches often deliver better performance, but they are computationally expensive and time-consuming.

Our proposed GraphIC involves using LLM-generated output to select ICEs, similar to Skill-kNN and DQ-LoRe. Although methods increase retrieval time due to invoking the LLM during the retrieval, their use of LLMs makes them suitable for more complex tasks.

\section{The Proposed GraphIC}

\subsection{Thought Graphs}\label{sec:thought_graph}

\begin{figure}[t]
    \centering
    \includegraphics[width=.9\linewidth]{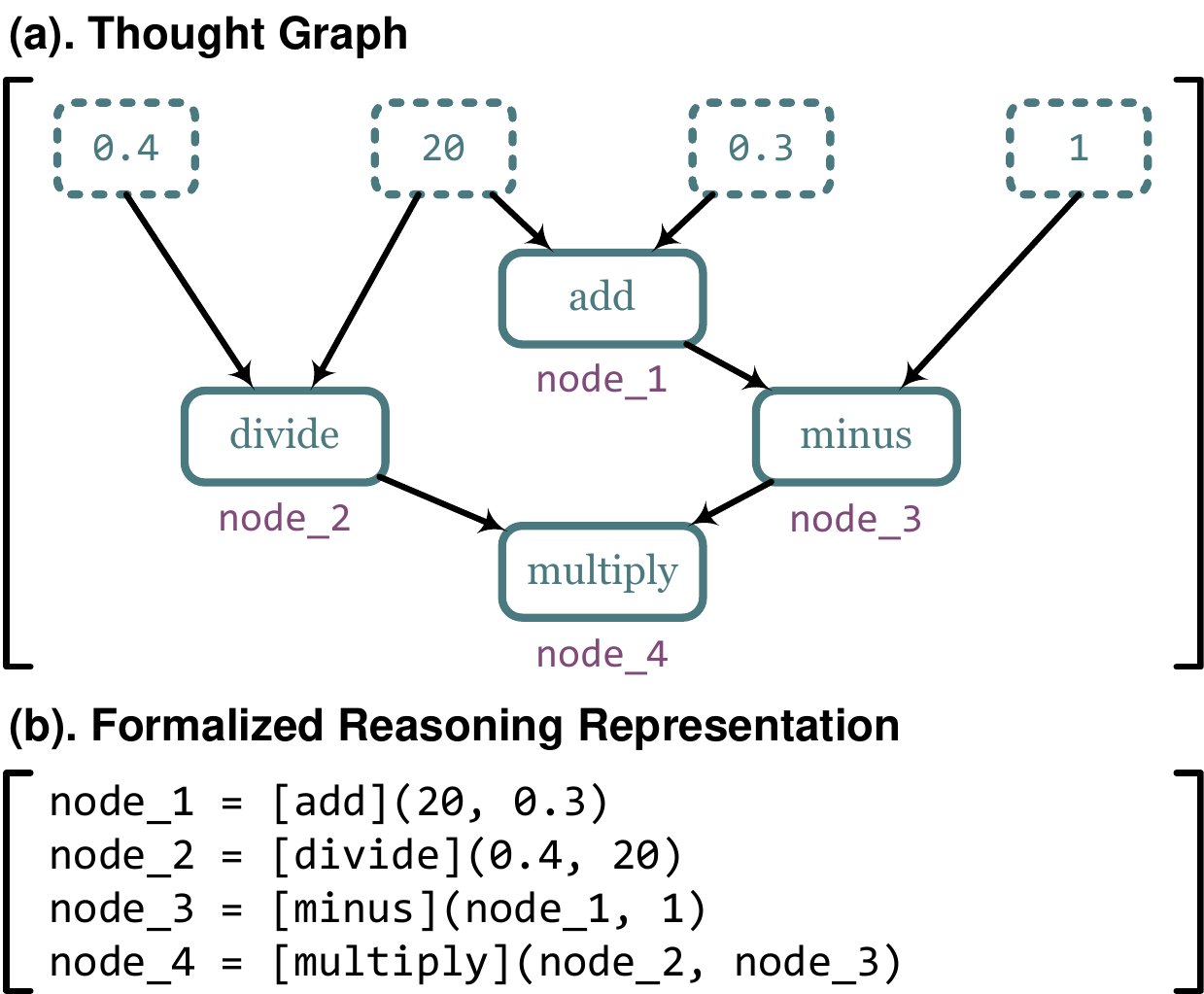}
    \caption{An example of a thought graph (a) and its corresponding FRR (b).}
    \label{fig:graph_frr}
\end{figure}

\textbf{Introduction of Thought Graphs.} As outlined earlier, thought graphs are directed, node-attributed graphs designed to model the reasoning process explicitly. In a thought graph, each node $i$ is attributed with an embedding vector \( x_i \in \mathbb{R}^{d} \) of a text, representing the key information of the current reasoning step. Directed edges between vertices represent the dependencies between steps. For example, if step B depends on the completion of step A, there is an edge from step A to step B.

As illustrated in Figure~\ref{fig:graph_frr} (a), the text corresponding to each vertex is "divide", "add", "minus", and "multiply". Therefore, the attributes of each vertex are $\mathrm{Emb}(\text{"divide"})$, $\mathrm{Emb}(\text{"add"})$, $\mathrm{Emb}(\text{"minus"})$, and $\mathrm{Emb}(\text{"multiply"})$, representing four base operations in a mathematical reasoning task. Here, $\mathrm{Emb}(\cdot)$ represents calculating the embedding of a given text using an embedding model, such as BERT. In this graph, an edge from the vertex labeled "add" to the vertex labeled "multiply" indicates that addition is performed first, and its result is then used in multiplication.

Due to the inherent differences between tasks, the text used to compute embedding varies slightly. Specifically, for mathematical reasoning, code generation, and logical reasoning tasks, we use mathematical operations, code snippets, and intermediate conclusions, respectively, as input texts. This is because, in mathematical reasoning tasks, a reasoning step can be represented by "which mathematical operations were performed on which variables"; in code generation tasks, it corresponds to "what code was executed"; and in logical reasoning tasks, it reflects "which existing conclusions were used to derive new conclusions". 
Examples of thought graphs for the four benchmark datasets are provided in the supplementary material.

\textbf{Construction of Thought Graphs.} For candidate examples, which include ground-truth natural language reasoning processes (i.e., chain-of-thought answers), we employ the following method to generate thought graphs. First, we use an LLM to formalize the reasoning process for each example. This formalized representation is called the Formalized Reasoning Representation (FRR). FRR filters irrelevant shallow semantics and extracts the key information from the reasoning process. Moreover, FRR is designed to be easily converted to a graph by applying rule-based parsing. The prompts used for generating FRRs and the pseudocode for parsing FRRs are shown in the supplementary material.

Figure~\ref{fig:graph_frr}~(b) illustrates the FRR of a thought graph. Each line in the FRR follows the format \verb|A = [B](C)|, where \verb|A| represents the name of the vertex, denoting the vertex itself. \verb|B| represents the label of the vertex, which will be used to compute its embedding. \verb|C| denotes the parent vertex of the given vertex. For example, in the FRR shown in Figure~\ref{fig:graph_frr}~(b), by parsing the first line, we create \verb|node_1|, set its label to "add", and add edges from "20" and "0.3" to \verb|node_1|. The remaining lines are parsed similarly to the first one. Note that in Figure~\ref{fig:graph_frr}~(a), the vertices representing numbers are indicated by dashed lines, which signifies that these vertices do not represent specific operations and will be removed once the entire graph is constructed.

For queries, since they lack ground-truth reasoning processes, we first use an LLM to generate pseudo natural language reasoning processes. We then follow the same procedure as with candidate examples to create thought graphs. Note that it is unnecessary to ensure the correctness of the generated reasoning process, as it is only used to aid in the retrieval of ICEs and is not part of the final answer. In fact, when a generated answer is incorrect, it is often not entirely wrong, but partially correct, which still reflects the reasoning process and can be helpful for retrieving examples. This ensures that GraphIC is robust to inaccuracies in the generated reasoning process. Even when the reasoning process is incorrect, GraphIC still maintains high accuracy. A detailed discussion and related experiments are provided in the supplementary material.

\subsection{Similarity Measure for Thought Graphs}\label{sec:similarity}

In this subsection, we first introduce a simplified case to illustrate our core idea, then extend it to propose our similarity measure.

\textbf{Simplified Case.} Consider a thought graph \(G\) with vertices \( v_1, \dots, v_n \), where each vertex \( v_i \) corresponds to a reasoning step with an attribute vector \(x_i \in \mathbb{R}^d\) (e.g., an embedding of a mathematical operation). Let \( X = [x_1, \dots, x_n] \in \mathbb{R}^{n \times d} \) denote the matrix of concatenated attributes. The adjacency matrix \( A \) of \( G \) is defined such that \( A_{ij} = 1 \) if there is a directed edge from \( v_i \) to \( v_j \), and \( A_{ij} = 0 \) otherwise. For a vertex \(v_i\), let \(\text{pa}(v_i)\) denote the set of its parent vertices (direct predecessors).

Generally, given two thought graphs $G_1$ and $G_2$, the similarity \(s(G_1, G_2)\) is computed in two steps: (1). \textbf{Extract Reasoning Pattern:} Solve Equation~(\ref{eqn:w_def1}) for \(G_1\) to obtain \(W_1\), which encapsulates \(G_1\)’s reasoning pattern. (2). \textbf{Compute Applicability:} Evaluate how well the reasoning pattern \(W_1\) applies to thought graph \(G_2\). This applicability, denoted as \(\mathcal{A}(W_1; G_2)\), serves as the similarity measure \(s(G_1, G_2)\). We now provide a detailed explanation of these two steps:

Firstly, we derive a matrix \(W\) that captures directional relationships in \(G\). This is achieved by solving the following optimization problem:
\begin{equation}\label{eqn:w_def1}
\begin{aligned}
    W &= \mathop{\arg\max}\limits_{\substack{W \in \mathbb{R}^{d \times d}, \\ \|W\|_F=1}} \mathcal{A}(W; G), \quad \text{where} \\
    \mathcal{A}(W; G) &= \sum_{i=1}^n \underbrace{(\sum_{j \in \text{pa}(v_i)} x_j)^\top}_{z_i} W x_i. \\
\end{aligned}
\end{equation}
Here, the term \(z_i = \sum_{j \in \text{pa}(v_i)} x_j \) calculates the sum of the attributes of \( v_i \)’s parent nodes, aggregating precursor information for step \( v_i \). $z_i^\top W$ represents a linear transformation of this information. $z_i^\top W x_i$ measures the inner product similarity between transformed precursor information \(W z_i\) and the current step’s attribute \( x_i \). By maximizing \( \mathcal{A}(W; G) \) under the Frobenius norm constraint (\(\|W\|_F = 1\)), \( W \) learns to predict the attribute of the next step based on precursor information, effectively encoding the directional relationships in \( G \).

For instance, consider a thought graph \( G \) with two vertices: \( v_1 \) with attribute \( x_1 = \mathrm{Emb(\text{"add"})} \) and \( v_2 \) with \( x_2 = \mathrm{Emb(\text{"multiply"})} \), connected by an edge \( v_1 \to v_2 \). In this setup, \( W \) learns to map \( x_1 \) ("add") to \( x_2 \) ("multiply"), indicating that "multiply" typically follows "add" in \( G \). When this learned mapping is applied to another graph \( G' \), a high value of \( \mathcal{A}(W; G') \) suggests that \( G' \) also contains frequent "add" $\to$ "multiply" sequences, reflecting similar problem-solving logic. Therefore, \( \mathcal{A}(W; G') \) quantifies how well the reasoning pattern of \( G \) applies to \( G' \). A higher \( \mathcal{A}(W; G') \) indicates an alignment between the two graphs' transitions, indicating that \( G' \) follows a similar problem-solving trajectory. This metric, therefore, provides an effective measure of the similarity in reasoning between \( G \) and \( G' \), making it suitable for retrieving ICEs that share compatible problem-solving logic.

\textbf{Proposed Similarity Metric.} The similarity metric in GraphIC builds upon the simplified case while introducing two critical enhancements to better model reasoning dynamics and computational efficiency.

Firstly, in Equation~(\ref{eqn:w_def1}), the information state $z_i$ for each node aggregates only its direct parent attributes. However, human reasoning often incorporates information from multiple prior steps and occasionally revisits initial premises. To capture this, we propose an iterative aggregation process:
\begin{equation}\label{eqn:iter}
    Z^{(h+1)} = [(1-\lambda)\tilde{A} + \lambda \tilde{B} + I]Z^{(h)},
\end{equation}
where $Z^{(h)}$ represents node information states at iteration $h$; $Z^{(0)}$ is initialized with node attributes $X$; $I$ is identity matrix; $\tilde{A} = D_A^{-\frac{1}{2}}AD_A^{-\frac{1}{2}}$, $D_A = \mathrm{diag}(\mathrm{deg}_{\mathrm{out}}(v_1), \dots, \mathrm{deg}_{\mathrm{out}}(v_1))$; $\tilde{B} = D_B^{-\frac{1}{2}}BD_B^{-\frac{1}{2}}$, and $B$ is traceback matrix enabling backtracking to root nodes, defined as $B_{ij} = 1$ if $\text{deg}_{\text{in}}(v_j) = 0$ and $\text{deg}_{\text{in}}(v_i) > 0$, otherwise $B_{ij} = 0$, with $\text{deg}_{\text{in}}(v_j)$ representing the in-degree of node $v_j$. A visual example of matrix $B$ is presented in the supplementary material. $\lambda$ is a balancing hyperparameter ($0\leq \lambda \leq 1$).

In each iteration, $z_i$ is updated by combining three information sources, corresponding to the matrices $A$, $B$, and $I$. Specifically, $AZ^{(h)}$ aggregates information from direct parent nodes, similar to the simplified case; $BZ^{(h)}$ facilitates backtracking in the reasoning process; and $IZ^{(h)}$ maintains current information.

Secondly, we observe that directly solving the optimization problem in Equation~(\ref{eqn:w_def1}) presents challenges related to both uniqueness and computational efficiency. Specifically, the number of vertices \( n \) in the thought graph is typically much smaller than the embedding dimension \( d \) (i.e., \( n \ll d \)). For instance, in the GSM8K dataset, thought graphs often contain fewer than 10 vertices, whereas the embedding dimension \( d \) can reach up to 768 when using BERT. This large disparity in dimensions leads to a non-unique solution for \( W \). Additionally, storing and computing a \( d \times d \) matrix is computationally expensive. To address both the issues of uniqueness and computational efficiency, we constrain \( W \) to be rank 1, allowing it to be expressed as \( W = \alpha \beta^\top \). As a result, \( \mathcal{A}(W; G) \) simplifies to:
\begin{equation}
    \mathcal{A}(W; G) = \mathcal{A}(\alpha, \beta; G) = \sum_{i=1}^{n} z_i^{\top}\alpha\beta^{\top}x_i
.\end{equation}
Both theoretical analysis and empirical evidence, presented in the supplementary material, demonstrate that the rank-1 assumption has minimal impact on solving the optimization problem.

Furthermore, with the rank-1 assumption, we can obtain the closed-form solution of Equation~(\ref{eqn:w_def1}), as shown in Equation~(\ref{eqn:rank_1_sol}), which greatly reduces the computational complexity of solving the optimization problem. The proof is shown in the supplementary material
\begin{equation}\label{eqn:rank_1_sol}
    \begin{aligned}
        W = \alpha\beta^\top, \alpha = U[0,:], \beta = V[0,:] \\
        \text{where}\quad U,\Sigma, V = \mathrm{SVD}(X^\top Z).
    \end{aligned}
\end{equation}
\textbf{Discussion.} Here, we compare proposed similarity with a widely used attributed-graph similarity to highlight two key properties of our metric: encoding directional relationships and asymmetry.

A common attributed-graph similarity metric computes the graph's embedding through an iterative formula, and then evaluates the similarity between graphs by calculating the cosine similarity of their embeddings. The iterative formula generally takes the following form:
\begin{equation}\label{eqn:emb_sim}
    X^{(h+1)} = \tilde{A}X^{(h)},\ X^{(0)}=X
.\end{equation}

Firstly, the embedding-based approach fails to effectively capture the sequential relationships between vertices, which are crucial in multi-step reasoning tasks. In these tasks, the order of operations can significantly alter the reasoning pattern. Therefore, the embedding-based similarity is not well-suited for such tasks. In contrast, our proposed similarity accounts for the transformation of information from previous vertices to current ones, inherently capturing these sequential relationships.

Secondly, our proposed similarity is asymmetric. Specifically, for two thought graphs $G_1$ and $G_2$, $s(G_1, G_2) \ne s(G_2, G_1)$. This asymmetry reflects real-world scenarios. For instance, as illustrated in the supplementary material, mathematical problem A might be a subproblem of problem B. In such a case, referencing B can help resolve A, but referencing A does not necessarily resolve B.

\subsection{Example Retrieval}

After introducing thought graphs and the proposed similarity metric, we now describe how GraphIC enhances ICL. During the preparation stage, GraphIC generates thought graphs for all candidate examples and estimates their reasoning patterns, denoted as $W_1, \dots, W_N$, where $N$ is the number of candidates. Then, given a query, GraphIC creates a thought graph for the query, denoted as $G^q$, computes the applicability of each reasoning pattern to this thought graph, $\mathcal A(W_i, G^q)$, and then selects the top $k$ as ICEs.

\section{Experiments}
\subsection{Experimental Setup}
\textbf{Datasets and Evaluations.} We conduct experiments across four multi-step reasoning tasks: mathematical reasoning (GSM8K~\cite{cobbe2021training} and AQUA~\cite{ling2017program}), code generation (MBPP~\cite{austin2021program}), and logical reasoning (ProofWriter~\cite{tafjord-etal-2021-proofwriter}). Model performance is measured by answer accuracy for GSM8K, AQUA, and ProofWriter, and by the pass@1 metric~\cite{chen2021evaluating} for MBPP.

\textbf{Models and Hyper-parameters.} We use both an open-source and a closed-source model. For the open-source model, we select Llama-3.1-8B-Instruct (hereafter referred to as Llama-3), one of the most advanced 7B-level models. For the closed-source model, we choose GPT-4o-mini, balancing performance and testing costs. Unless explicitly mentioned otherwise, all evaluations use an 8-shot setting, which is the most common setting in chain-of-thought scenarios. We set the temperature to 1e-5 to minimize randomness, ensuring consistency across generations, and use 3 iterations in Equation~(\ref{eqn:iter}) with \( \lambda \) values from \( \{0, 0.1, 0.2, 0.3\} \) (see the supplementary material for specific details). More experiment details are in the supplementary material.

\subsection{Baselines}
We compare GraphIC against six training-free retrieval methods spanning random, similarity-based, diversity-based, and complexity-based approaches, including:
\begin{mylist}
    \item \textbf{Random} randomly selects $k$ unique ICEs from the candidate set; 
    \item \textbf{BM25}~\cite{robertson2009probabilistic} selects the top $k$ examples based on BM25 scoring;
    \item \textbf{BERT}~\cite{devlin-etal-2019-bert} is a dense retriever using cosine similarity with BERT-base-uncased embeddings;
    \item \textbf{Complex-CoT}~\cite{fu2022complexity} selects $k$ examples based on complexity, quantified by newline characters;
    \item \textbf{Auto-CoT}~\cite{zhangautomatic} clusters candidates and selects the closests to each cluster center; and 
    \item \textbf{Skill-kNN}~\cite{an2023skill} prompts LLM to generate task-relevant skills for query and candidates, followed by dense retrieval.
\end{mylist}

We also compare with four training-based methods, which encompass both single-example and combination-example retrieval strategies, including:
\begin{mylist}
    \item \textbf{EPR}~\cite{rubin2022learning} is trained to retrieve the single most relevant ICE, with top $k$ examples being selected during inference; 
    \item \textbf{CEIL}~\cite{ye2023compositional} uses determinantal point processes to select ICEs balancing similarity and diversity; 
    \item \textbf{DQ-LoRe}~\cite{xiong2024dqlore} uses dual queries and low-rank approximation re-ranking to identify ICEs; and
    \item \textbf{GistScore}~\cite{guptagistscore} encodes task-specific information into gist tokens for selecting ICEs.
\end{mylist}

\subsection{Main Results}

Table~\ref{tab:main_results} shows the performance comparison results. 
As a training-free method, GraphIC consistently outperforms both training-free and training-based baselines in most settings.

We find that training-based baselines generally outperform training-free baselines, as they learn explicit patterns from in-context examples. For instance, with Llama-3, training-based methods average over 67\% performance, while the top training-free baseline (Complex-CoT) reaches only 66.54\%. Among the training-based baselines, DQ-LoRe is the most effective, reaching 68.33\%. This method leverages LLM-generated outputs during retrieval, demonstrating that incorporating LLM outputs improves performance on reasoning tasks. Among training-free baselines, Complex-CoT and BM25 perform best. Notably, both use asymmetric retrieval, reinforcing that asymmetric retrieval aligns well with real-world reasoning scenarios.

GraphIC integrates the strengths of existing methods—utilizing LLM outputs for reasoning tasks and asymmetric retrieval—while introducing a reasoning-aware representation and similarity metric. Consequently, it not only surpasses all training-free baselines by (2.57\% with GPT-4o-mini and 4.29\% with Llama-3), but also outperforms all training-based methods (by 1.18\% using GPT-4o-mini and 2.5\% using Llama-3), highlighting its effectiveness in reasoning tasks.

A closer analysis highlights GraphIC’s substantial advantages in mathematical and logical reasoning tasks compared to code generation, particularly for complex problem instances. For example, on GSM8K, GraphIC outperforms all baselines by 0.65\% and 3.57\% with the two LLMs. In the more challenging AQUA dataset, the performance improvements become even more pronounced, reaching 3.47\% and 7.64\%.

\begin{table*}[ht]
    \centering
    \setlength{\tabcolsep}{3mm}{
        \begin{tabular}{ccccccc}
            \toprule
            LLM & Model & GSM8K & AQUA & MBPP & ProofWriter & Avg.\\
            \midrule
            \multirow{12}{*}{GPT-4o-mini}
                                         & Random & 92.90 (0.31) & 71.58 (0.72) & 72.76 (0.74) & 64.90 (0.93) & 75.54 (0.36) \\
                                         & BM25 & 92.64 & 70.47 & 73.4 & 66.25 & 75.69 \\
                                         & BERT & 93.02 & 66.93 & 74.2 & 65.25 & 74.85 \\
                                         & Complex-CoT & 92.49 & 67.32 & 74.2 & 64.25 & 74.57 \\
                                         & Auto-CoT & 92.72 & 69.69 & 73.8 & 62.25 & 74.62 \\
                                         & Skill-kNN & 92.34 & 71.65 & 72.0 & 66.00 & 75.50 \\
                                         \cmidrule[0.2pt]{2-7}
                                         & EPR & 93.02 & 72.04 & 73.8 & 68.50 & 76.84 \\
                                         & CEIL & 92.57 & \underline{72.44} & 73.8 & \underline{69.50} & \underline{77.08} \\
                                         & DQ-LoRe & \underline{93.32} & 69.69 & \underline{74.6} & 66.50 & 76.03 \\
                                         & GistScore & 93.25 & 69.69 & 72.8 & 67.00 & 75.69 \\
                                         \cmidrule[0.2pt]{2-7}
                                         & GraphIC & \textbf{93.48} & \textbf{73.62} & \textbf{75.2} & \textbf{70.75} & \textbf{78.26} \\
            \midrule
            \multirow{12}{*}{\makecell{Llama-3.1\\-8B-Instruct}} 
                                         & Random & 78.86 (0.87) & 53.15 (1.85) & 57.72 (1.06) & 76.10 (2.45) & 66.46 (0.84)\\
                                         & BM25 & 77.71 & 46.85 & \underline{62.0} & 77.75 & 66.08 \\
                                         & BERT & 74.15 & 50.39 & 60.8 & 73.75 & 64.77 \\
                                         & Complex-CoT & \underline{79.30} & 50.00 & 58.6 & 78.25 & 66.54 \\
                                         & Auto-CoT & 72.78 & 42.91 & 58.4 & 78.00 & 63.02 \\
                                         & Skill-kNN & 77.56 & 50.39 & 60.8 & 74.00 & 65.69 \\
                                         \cmidrule[0.2pt]{2-7}
                                         & EPR & 75.66 & 53.94 & \underline{62.0} & 79.25 & 67.71 \\
                                         & CEIL & 75.51 & 51.97 & \textbf{62.4} & 81.00 & 67.72 \\
                                         & DQ-LoRe & 77.93 & \underline{54.33} & 59.8 & \underline{81.25} & \underline{68.33}\\
                                         & GistScore & 74.60 & 44.49 & 60.4 & 79.50 & 64.75 \\
                                         \cmidrule[0.2pt]{2-7}
                                         & GraphIC & \textbf{79.98} & \textbf{57.48} & 61.6 & \textbf{84.25} & \textbf{70.83} \\
            \bottomrule
        \end{tabular}
    }
    \caption{Main results on two LLMs and four datasets. For random retrieval, we present the mean and standard deviation derived from five independent experiments.}
    \label{tab:main_results}
\end{table*}

\subsection{Ablation Study} We conduct a series of ablation studies to systematically evaluate the contribution of each component in GraphIC, focusing on three main aspects: filtering shallow semantic information, integrating the graph structure, and the proposed similarity metric.

To this end, we develop several variants of the GraphIC model:
\begin{mylist}
    \item \textbf{Text} relies solely on text embedding, the same as the BERT approach;
    \item \textbf{FRR} retrieves examples using text embeddings of FRRs;
    \item \textbf{Graph} employs Equation~(\ref{eqn:emb_sim}) to generate graph embeddings and use cosine similarity to retrieve examples;
    \item \textbf{Graph+B} employs Equation~(\ref{eqn:iter}) to generate graph embeddings and use cosine similarity to retrieve examples;
    \item \textbf{Graph+S} excludes the backtracking mechanism from the full GraphIC model during computing $Z$;
    and 
    \item \textbf{Graph+B+S} represents the full GraphIC model, integrating all components.
\end{mylist}

We conduct experiments using Llama-3 across four datasets, with the results presented in Table~\ref{tab:ablation}. First, we observe that employing FRR to compute text embeddings significantly improves performance (FRR versus Text), especially on GSM8K and ProofWriter, with performance increasing from 74.15 to 78.31 and from 73.75 to 82.50, respectively. This improvement is due to FRR's ability to filter out substantial shallow semantic noise. Second, converting FRR into a graph representation (i.e., a thought graph) further improves performance (Graph versus FRR), as seen on the AQUA dataset, where performance rises from 50.78 to 54.72. This suggests that the graph structure is better suited for capturing the reasoning process than linear text. Third, introducing the reasoning-aware similarity metric also enhances performance (Graph+B+S versus Graph), increasing from 54.72 to 57.48 on AQUA, highlighting the effectiveness of reasoning-aware similarity. Additionally, our experiments show that incorporating the traceback mechanism better simulates the thought process, leading to further performance gains (Graph+B versus Graph, and Graph+B+S versus Graph+S).

In summary, the findings emphasize that each component of GraphIC contributes significantly to improving model performance, with the most substantial gains occurring when all components are combined.

\begin{table}[htbp]
    \centering
    \setlength{\tabcolsep}{1mm}{
        \begin{tabular}{lcccc}
            \toprule
            Model & GSM8K & AQUA & MBPP & ProofWriter \\
            \midrule
            Text & 74.15 & 50.39 & 60.8 & 73.75 \\
            FRR & 78.31 & 50.78 & 60.4 & 82.50 \\
            Graph & 78.46 & 54.72 & 60.4 & 83.50 \\
            \quad+B & 78.92 & 56.30 & 61.0 & 83.75 \\
            \quad+S & 79.07 & 49.21 & 60.4 & \textbf{84.25} \\
            \quad+B+S & \textbf{79.98} & \textbf{57.48} & \textbf{61.6} & \textbf{84.25} \\
            \bottomrule
        \end{tabular}
    }
    \caption{Ablation Study.}
    \label{tab:ablation}
\end{table}

\subsection{Analysis}

\begin{figure*}[ht]
    \centering
    \subfloat[GSM8K]{
    \includegraphics[width=0.2\linewidth]{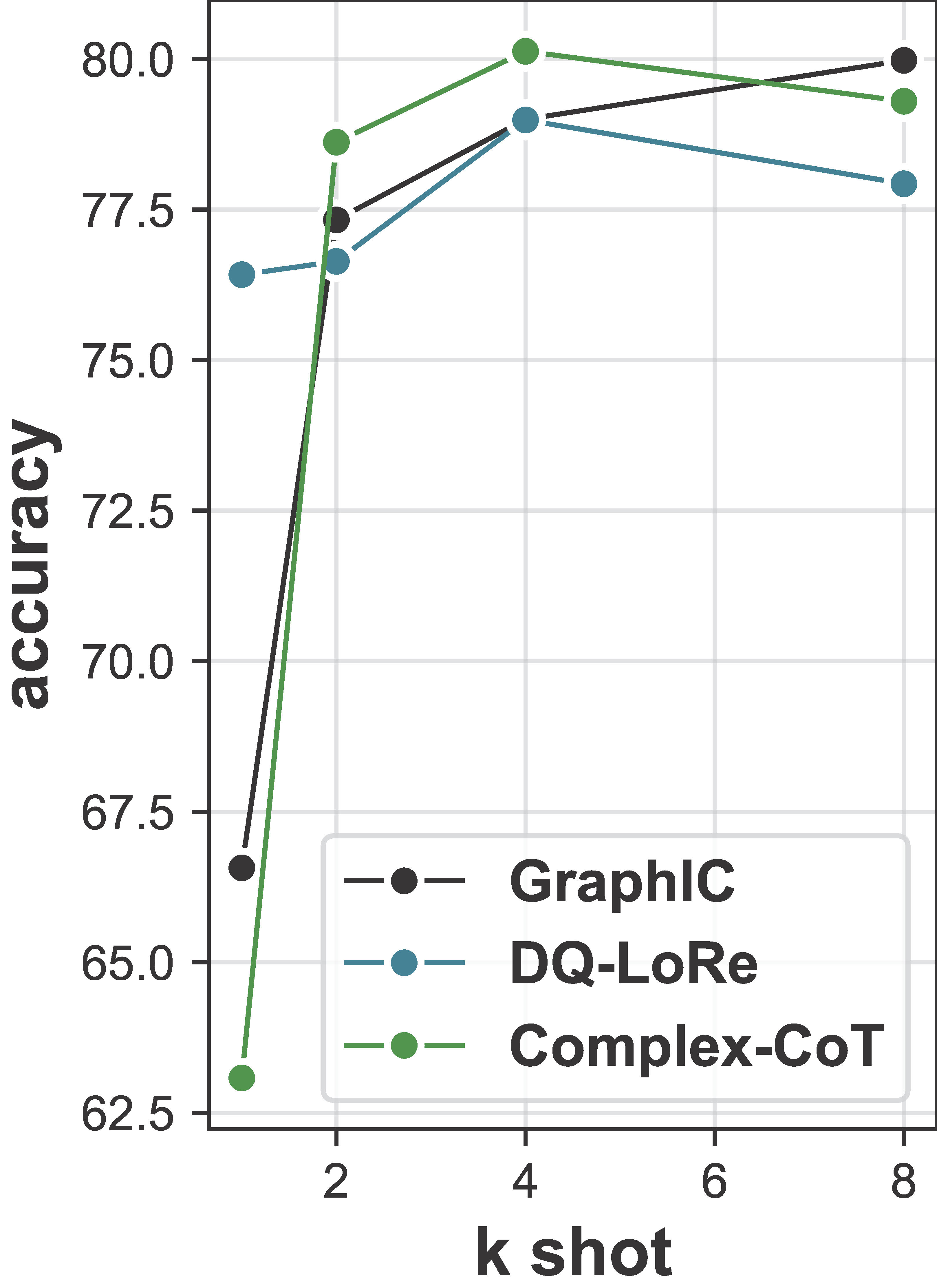}
    }\hspace{0.02\linewidth}
    \subfloat[AQUA]{
    \includegraphics[width=0.2\linewidth]{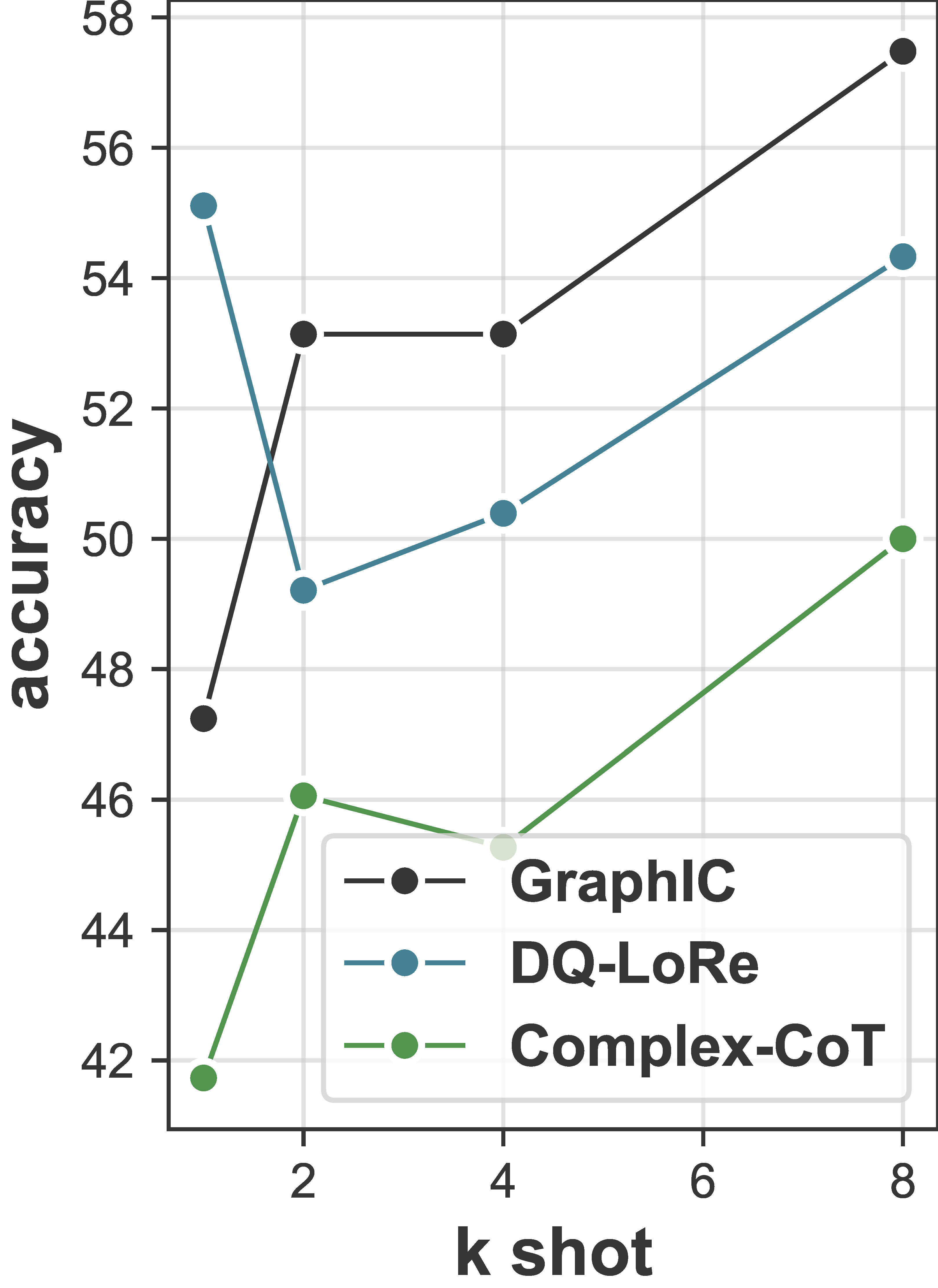}
    }\hspace{0.02\linewidth}
    \subfloat[MBPP]{
    \includegraphics[width=0.2\linewidth]{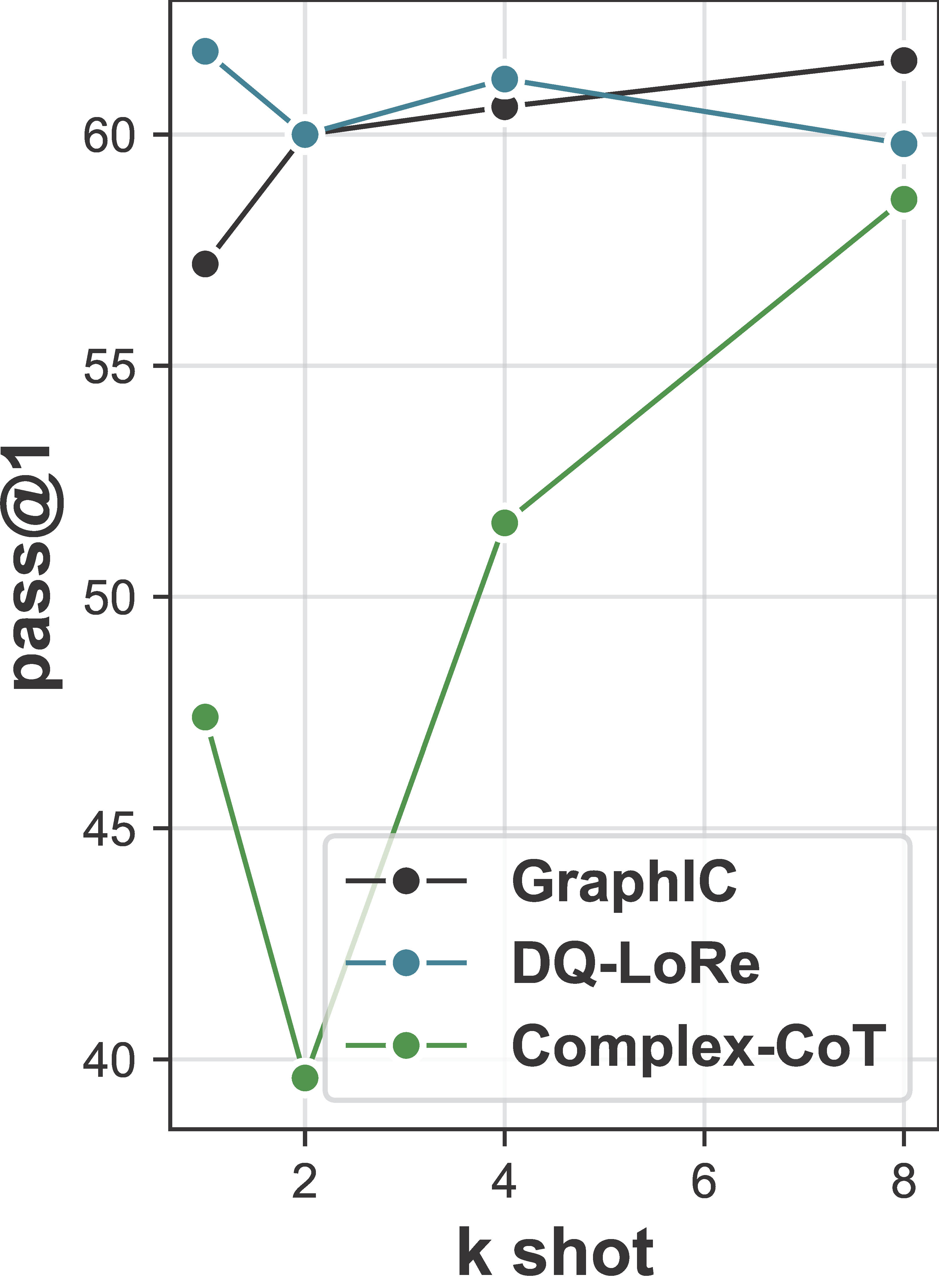}
    }\hspace{0.02\linewidth}
    \subfloat[ProofWriter]{
    \includegraphics[width=0.2\linewidth]{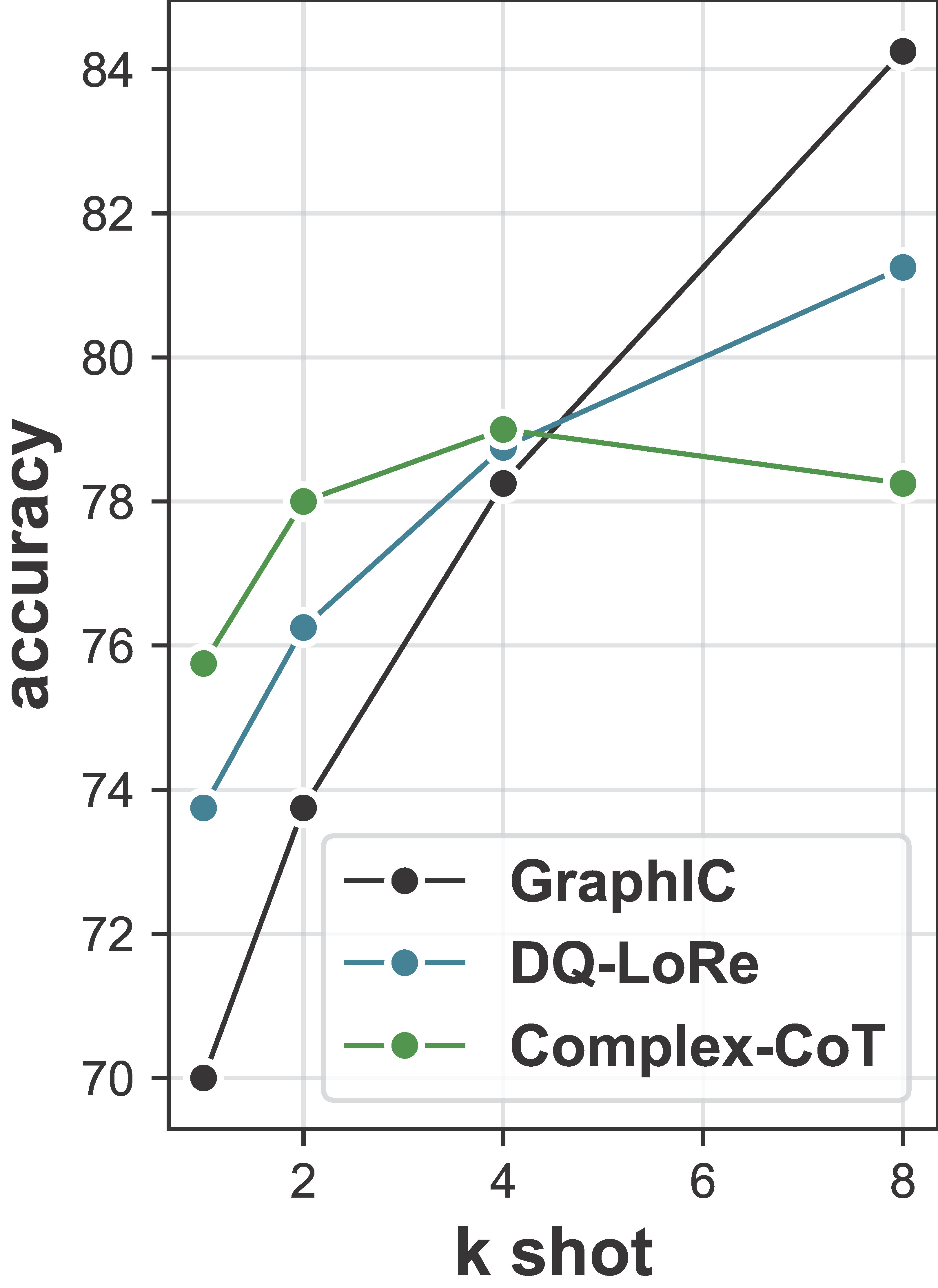}
    }
    \caption{Performance of GraphIC and Top Training-based/Training-free Baselines (DQ-LoRe and Complex-CoT) across 1–8 Shot Settings.}
    \label{fig:k_shot}
\end{figure*}

\begin{figure*}[thbp]
    \centering
    \subfloat[Ground-Truth]{
    \includegraphics[width=0.2\linewidth]{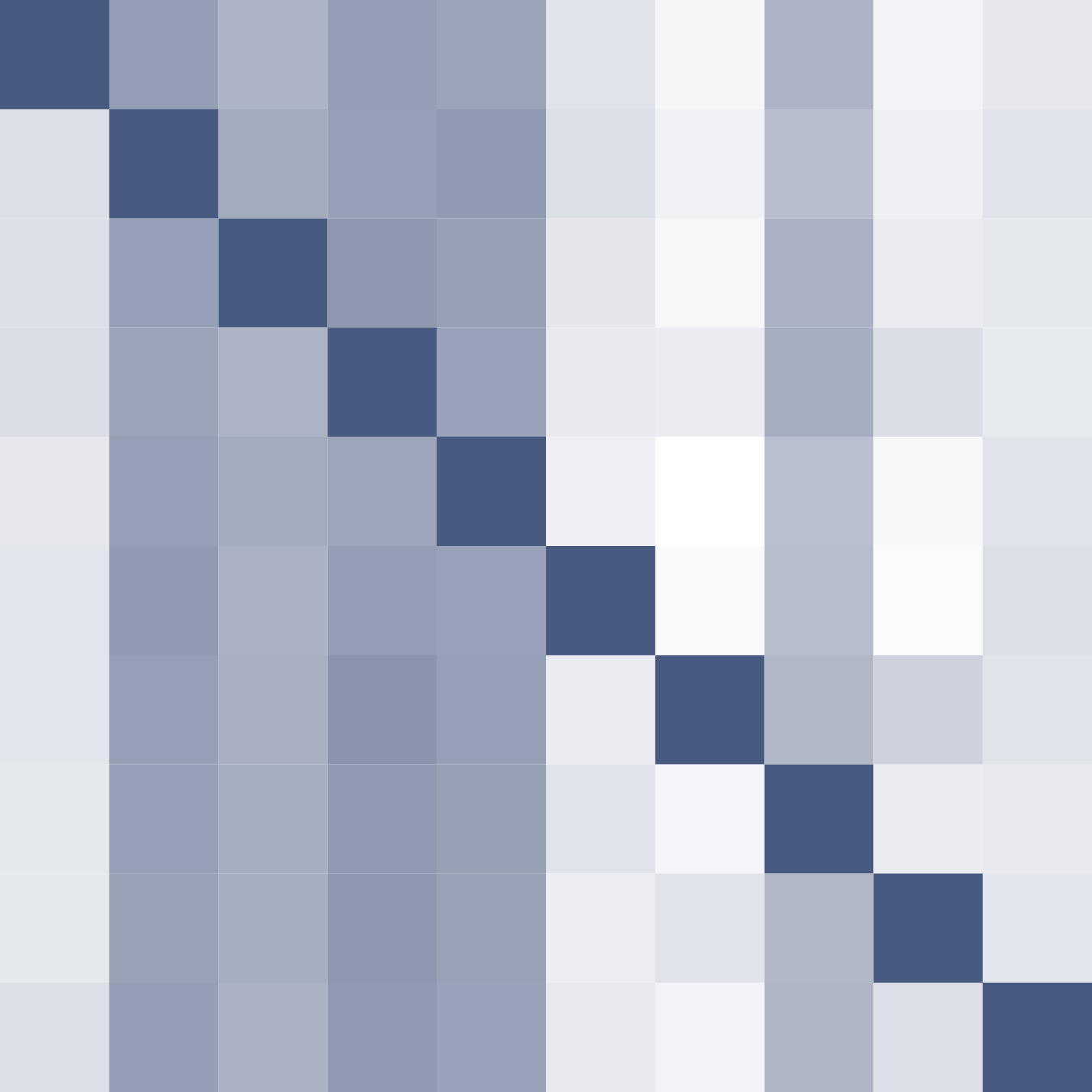}
    }\hspace{0.02\linewidth}
    \subfloat[Complex-CoT]{
    \includegraphics[width=0.2\linewidth]{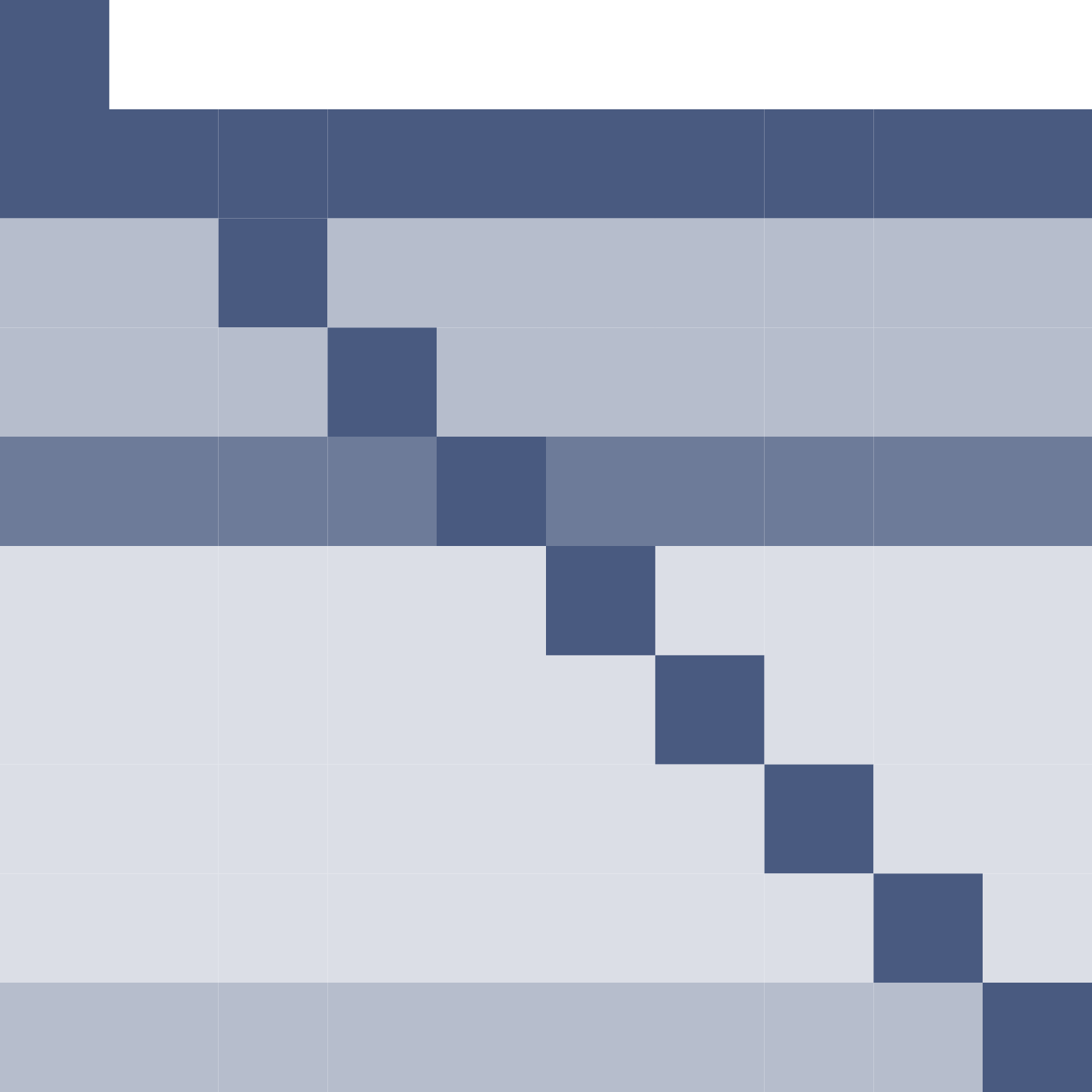}
    }\hspace{0.02\linewidth}
    \subfloat[DQ-LoRe]{
    \includegraphics[width=0.2\linewidth]{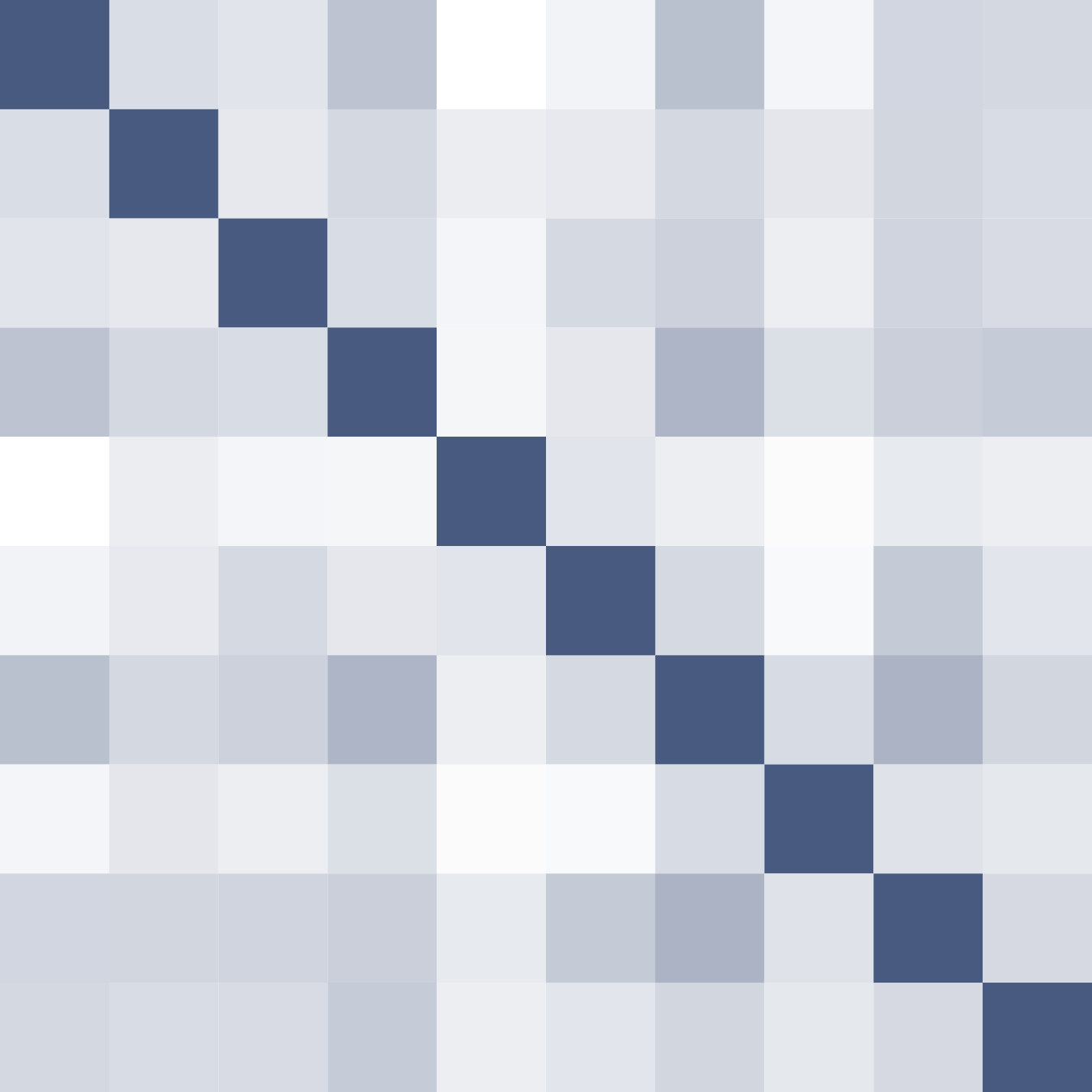}
    }\hspace{0.02\linewidth}
    \subfloat[GraphIC]{
    \includegraphics[width=0.2\linewidth]{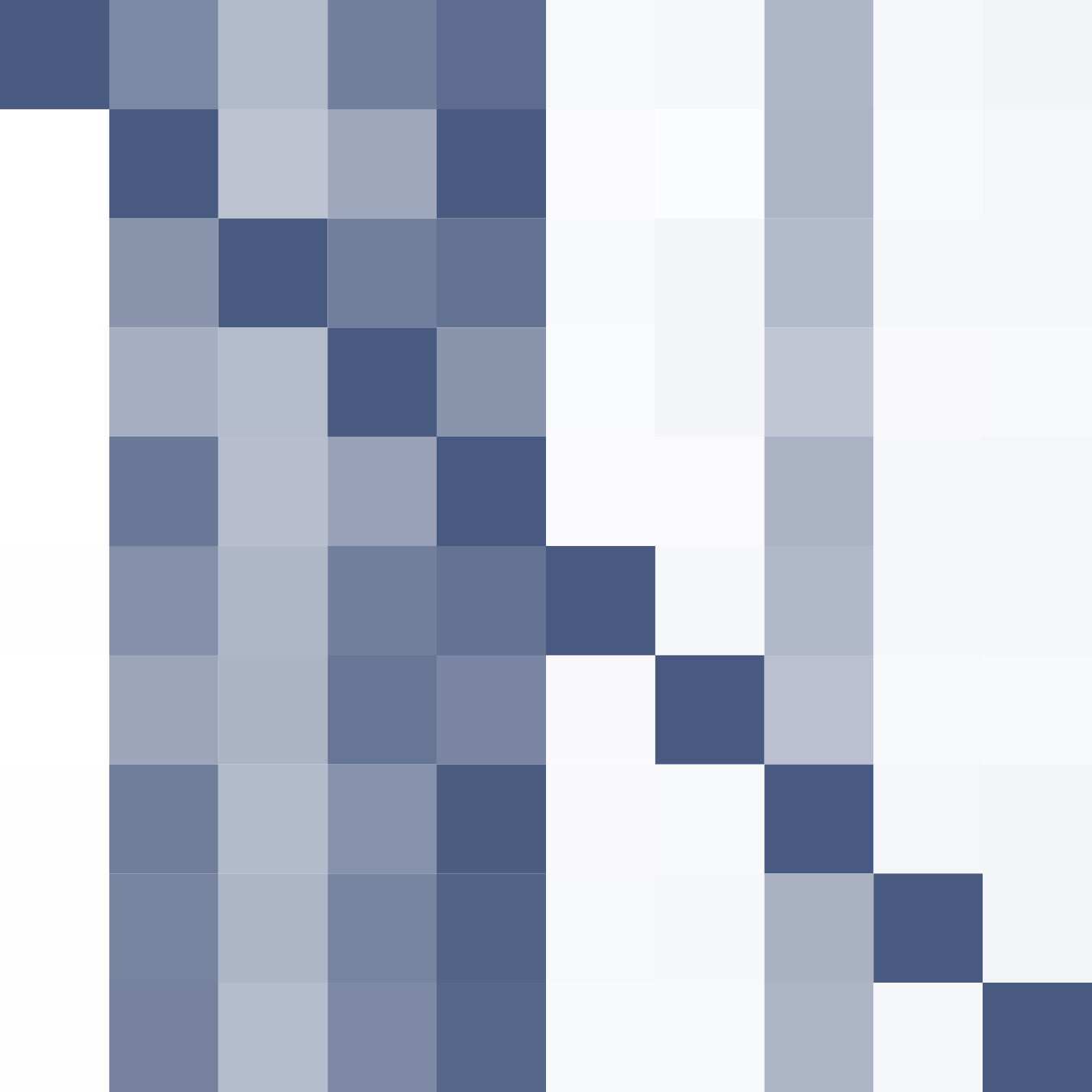}
    }
    \caption{Ground-truth matrix and score matrices of various models. The matrix values have been linearly scaled to the range [0,1], with darker shades representing values closer to 1, and the diagonal elements have been set to 1.}
    \label{fig:symmetry}
\end{figure*}

\textbf{The performance of GraphIC steadily improves as $k$ increases.} We analyze how the performance of GraphIC and top training-based/free baselines (DQ-LoRe and Complex-CoT) changes as $k$ increases. Specifically, we vary the number of examples $k$ in the set \{1, 2, 4, 8\} and use Llama-3 as LLM. 

As shown in Figure~\ref{fig:k_shot}, we observe that model performance generally improves with $k$. Notably, GraphIC consistently performs better as $k$ increases, whereas other methods experience performance degradation. Additionally, training-based and training-free methods show different trends with $k$ increases. Specifically, training-based methods, such as DQ-LoRe, directly optimize the probability of LLM generating correct answers in 1-shot scenarios. As a result, they tend to achieve superior performance in low-shot settings, especially in 1-shot cases. However, as $k$ increases, the performance gains of these methods tend to slow down or even decline. In contrast, training-free methods, like Complex-CoT and GraphIC, typically underperform compared to training-based methods in low-shot settings. However, as the value of \( k \) increases, they exhibit more stable improvement.

\textbf{Asymmetric retrieval aligns more closely with real-world scenarios.} As previously discussed, GraphIC is an asymmetric method that reflects real-world scenarios. To further validate this asymmetry, we randomly select 10 examples from GSM8K and compute their "improve matrix" \( S \), which captures the mutual improvement relationships between examples. In this matrix, \( S_{ij} \) is defined as the probability of solving the \( j \)-th example correctly when using the \( i \)-th example as the ICE. The resulting improve matrix (Figure~\ref{fig:symmetry}~(a)), clearly illustrates the asymmetry in real-world scenarios: while the \( i \)-th example may aid in solving the \( j \)-th example accurately, the reverse is not necessarily true.

Additionally, we examine the alignment between the score matrices of GraphIC and top training-free/based models with the ground-truth improve matrix. Existing ICE retrieval methods select ICEs by assigning scores to (candidate, query) pairs and comparing them. Similar to the improve matrix, we can compute a score matrix for each method, reflecting the mutual improvement relationships between examples estimated by the method. Naturally, the closer a method’s score matrix aligns with the improvement matrix, the better it captures real-world relationships. The score matrices of Complex-CoT, DQ-LoRe, and GraphIC are shown in Figure~\ref{fig:symmetry}~(b), (c), and (d), respectively. The results clearly demonstrate that GraphIC aligns well with the ground-truth, while the other two models exhibit noticeable deviations.

\section{Conclusion}
We propose GraphIC, which employs a reasoning-aware representation—\textit{thought graph} and a similarity metric to retrieve ICEs. Our experiments show that GraphIC outperforms both training-free and training-based baselines, highlighting the importance of modeling reasoning structures and providing insights for future research on graph-based solutions for complex problem-solving.

\section*{Acknowledgments}

This work is supported by the National Science Foundation of China (62576091) and the Fundamental Research Funds for the Central Universities (2242025K30024). Additional support was provided by the Big Data Computing Center of Southeast University and the SEU Kunpeng \& Ascend Center of Cultivation. Y. Wang is sponsored by Beijing Nova Program.

\bibliography{aaai2026}

@InProceedings{xiong2024dqlore,
  author    = {Jing Xiong and Zixuan Li and Chuanyang Zheng and Zhijiang Guo and Yichun Yin and Enze Xie and Zhicheng YANG and Qingxing Cao and Haiming Wang and Xiongwei Han and Jing Tang and Chengming Li and Xiaodan Liang},
  booktitle = {The Twelfth International Conference on Learning Representations},
  title     = {{DQ-LoRe}: Dual Queries with Low Rank Approximation Re-ranking for In-Context Learning},
  year      = {2024},
}

@article{cobbe2021training,
  title={Training verifiers to solve math word problems},
  author={Cobbe, Karl and Kosaraju, Vineet and Bavarian, Mohammad and Chen, Mark and Jun, Heewoo and Kaiser, Lukasz and Plappert, Matthias and Tworek, Jerry and Hilton, Jacob and Nakano, Reiichiro and others},
  journal={arXiv preprint arXiv:2110.14168},
  year={2021}
}

@inproceedings{ling2017program,
  title={Program Induction by Rationale Generation: Learning to Solve and Explain Algebraic Word Problems},
  author={Ling, Wang and Yogatama, Dani and Dyer, Chris and Blunsom, Phil},
  booktitle={Findings of the Association for Computational Linguistics: ACL 2017},
  pages={158--167},
  year={2017}
}

@article{austin2021program,
  title={Program synthesis with large language models},
  author={Austin, Jacob and Odena, Augustus and Nye, Maxwell and Bosma, Maarten and Michalewski, Henryk and Dohan, David and Jiang, Ellen and Cai, Carrie and Terry, Michael and Le, Quoc and others},
  journal={arXiv preprint arXiv:2108.07732},
  year={2021}
}

@inproceedings{tafjord-etal-2021-proofwriter,
    title = "{P}roof{W}riter: Generating Implications, Proofs, and Abductive Statements over Natural Language",
    author = "Tafjord, Oyvind  and
      Dalvi, Bhavana  and
      Clark, Peter",
    editor = "Zong, Chengqing  and
      Xia, Fei  and
      Li, Wenjie  and
      Navigli, Roberto",
    booktitle = "Findings of the Association for Computational Linguistics: ACL-IJCNLP 2021",
    year = "2021",
    pages = "3621--3634",
}

@Article{robertson2009probabilistic,
  author    = {Robertson, Stephen and Zaragoza, Hugo and others},
  journal   = {Foundations and Trends{\textregistered} in Information Retrieval},
  title     = {The probabilistic relevance framework: {BM25} and beyond},
  year      = {2009},
  number    = {4},
  pages     = {333--389},
  volume    = {3},
}

@inproceedings{devlin-etal-2019-bert,
    title = "{BERT}: Pre-training of Deep Bidirectional Transformers for Language Understanding",
    author = "Devlin, Jacob  and
      Chang, Ming-Wei  and
      Lee, Kenton  and
      Toutanova, Kristina",
    editor = "Burstein, Jill  and
      Doran, Christy  and
      Solorio, Thamar",
    booktitle = "Proceedings of the 2019 Conference of the North {A}merican Chapter of the Association for Computational Linguistics: Human Language Technologies",
    year = "2019",
    pages = "4171--4186",
}

@inproceedings{fu2022complexity,
  title={Complexity-based prompting for multi-step reasoning},
  author={Fu, Yao and Peng, Hao and Sabharwal, Ashish and Clark, Peter and Khot, Tushar},
  booktitle={The Eleventh International Conference on Learning Representations},
  year = "2022"
}

@inproceedings{zhangautomatic,
  title={Automatic Chain of Thought Prompting in Large Language Models},
  author={Zhang, Zhuosheng and Zhang, Aston and Li, Mu and Smola, Alex},
  booktitle={The Eleventh International Conference on Learning Representations},
  year = "2022"
}

@inproceedings{an2023skill,
  title={Skill-Based Few-Shot Selection for In-Context Learning},
  author={An, Shengnan and Zhou, Bo and Lin, Zeqi and Fu, Qiang and Chen, Bei and Zheng, Nanning and Chen, Weizhu and Lou, Jian-Guang},
  booktitle={The 2023 Conference on Empirical Methods in Natural Language Processing},
  year = "2023"
}

@inproceedings{rubin2022learning,
  title={Learning To Retrieve Prompts for In-Context Learning},
  author={Rubin, Ohad and Herzig, Jonathan and Berant, Jonathan},
  booktitle={Proceedings of the 2022 Conference of the North American Chapter of the Association for Computational Linguistics: Human Language Technologies},
  pages={2655--2671},
  year={2022}
}

@inproceedings{ye2023compositional,
  title={Compositional Exemplars for In-context Learning},
  author={Ye, Jiacheng and Wu, Zhiyong and Feng, Jiangtao and Yu, Tao and Kong, Lingpeng},
  booktitle={Fortieth International Conference on Machine Learning},
  year={2023}
}

@InProceedings{guptagistscore,
  author    = {Gupta, Shivanshu and Rosenbaum, Clemens and Elenberg, Ethan R},
  booktitle = {Forty-First International Conference on Machine Learning},
  title     = {{GistScore}: Learning Better Representations for In-Context Example Selection with Gist Bottlenecks},
  year      = {2024},
}

@article{chen2021evaluating,
  title={Evaluating large language models trained on code},
  author={Chen, Mark and Tworek, Jerry and Jun, Heewoo and Yuan, Qiming and Pinto, Henrique Ponde De Oliveira and Kaplan, Jared and Edwards, Harri and Burda, Yuri and Joseph, Nicholas and Brockman, Greg and others},
  journal={arXiv preprint arXiv:2107.03374},
  year={2021}
}

@InProceedings{liu2022makes,
  author    = {Liu, Jiachang and Shen, Dinghan and Zhang, Yizhe and Dolan, William B and Carin, Lawrence and Chen, Weizhu},
  booktitle = {Proceedings of Deep Learning Inside Out (DeeLIO 2022): The 3rd Workshop on Knowledge Extraction and Integration for Deep Learning Architectures},
  title     = {What Makes Good In-Context Examples for {GPT-3}?},
  year      = {2022},
  pages     = {100--114},
}

@inproceedings{hu2022context,
  title={In-Context Learning for Few-Shot Dialogue State Tracking},
  author={Hu, Yushi and Lee, Chia-Hsuan and Xie, Tianbao and Yu, Tao and Smith, Noah A and Ostendorf, Mari},
  booktitle={The 2022 Conference on Empirical Methods in Natural Language Processing},
  pages={2627--2643},
  year={2022}
}

@inproceedings{agrawal2023context,
  title={In-context Examples Selection for Machine Translation},
  author={Agrawal, Sweta and Zhou, Chunting and Lewis, Mike and Zettlemoyer, Luke and Ghazvininejad, Marjan},
  booktitle={Findings of the Association for Computational Linguistics: ACL 2023},
  pages={8857--8873},
  year={2023}
}

@InProceedings{tonglet2023seer,
  author    = {Tonglet, Jonathan and Reusens, Manon and Borchert, Philipp and Baesens, Bart},
  booktitle = {The 2023 Conference on Empirical Methods in Natural Language Processing},
  title     = {{SEER}: A Knapsack approach to Exemplar Selection for In-Context {HybridQA}},
  year      = {2023},
  pages     = {13569--13583},
}

@inproceedings{cho2023prompt,
  title={Prompt-augmented linear probing: Scaling beyond the limit of few-shot in-context learners},
  author={Cho, Hyunsoo and Kim, Hyuhng Joon and Kim, Junyeob and Lee, Sang-Woo and Lee, Sang-goo and Yoo, Kang Min and Kim, Taeuk},
  booktitle={Proceedings of the AAAI Conference on Artificial Intelligence},
  pages={12709--12718},
  year={2023}
}

@inproceedings{gonen2023demystifying,
  title={Demystifying Prompts in Language Models via Perplexity Estimation},
  author={Gonen, Hila and Iyer, Srini and Blevins, Terra and Smith, Noah A and Zettlemoyer, Luke},
  booktitle={The 2023 Conference on Empirical Methods in Natural Language Processing},
  year={2023}
}

@inproceedings{wu2023self,
  title={Self-Adaptive In-Context Learning: An Information Compression Perspective for In-Context Example Selection and Ordering},
  author={Wu, Zhiyong and Wang, Yaoxiang and Ye, Jiacheng and Kong, Lingpeng},
  booktitle={Findings of the Association for Computational Linguistics: ACL 2023},
  year={2023}
}

@article{nguyen2023context,
  title={In-context example selection with influences},
  author={Nguyen, Tai and Wong, Eric},
  journal={arXiv preprint arXiv:2302.11042},
  year={2023}
}

@inproceedings{yang2023representative,
  title={Representative demonstration selection for in-context learning with two-stage determinantal point process},
  author={Yang, Zhao and Zhang, Yuanzhe and Sui, Dianbo and Liu, Cao and Zhao, Jun and Liu, Kang},
  booktitle={The 2023 Conference on Empirical Methods in Natural Language Processing},
  year={2023}
}

@inproceedings{li2023finding,
  title={Finding Support Examples for In-Context Learning},
  author={Li, Xiaonan and Qiu, Xipeng},
  booktitle={The 2023 Conference on Empirical Methods in Natural Language Processing},
  pages={6219--6235},
  year={2023}
}

@InProceedings{wang2024large,
  author    = {Wang, Xinyi and Zhu, Wanrong and Saxon, Michael and Steyvers, Mark and Wang, William Yang},
  booktitle = {Advances in Neural Information Processing Systems},
  title     = {Large language models are latent variable models: Explaining and finding good demonstrations for in-context learning},
  year      = {2023},
  pages     = {15614--15638},
  volume    = {36},
}

@inproceedings{zhang2022active,
  title={Active Example Selection for In-Context Learning},
  author={Zhang, Yiming and Feng, Shi and Tan, Chenhao},
  booktitle={The 2022 Conference on Empirical Methods in Natural Language Processing},
  pages={9134--9148},
  year={2022}
}

@Article{scarlatos2023reticl,
  author  = {Scarlatos, Alexander and Lan, Andrew},
  journal = {arXiv preprint arXiv:2305.14502},
  title   = {{RetICL}: Sequential retrieval of in-context examples with reinforcement learning},
  year    = {2023},
}

@inproceedings{lu2022dynamic,
  title={Dynamic Prompt Learning via Policy Gradient for Semi-structured Mathematical Reasoning},
  author={Lu, Pan and Qiu, Liang and Chang, Kai-Wei and Wu, Ying Nian and Zhu, Song-Chun and Rajpurohit, Tanmay and Clark, Peter and Kalyan, Ashwin},
  booktitle={The Eleventh International Conference on Learning Representations},
  year={2022}
}

@InProceedings{xureprompting,
  author    = {Xu, Weijia and Banburski, Andrzej and Jojic, Nebojsa},
  booktitle = {Forty-First International Conference on Machine Learning},
  title     = {Reprompting: Automated Chain-of-Thought Prompt Inference Through Gibbs Sampling},
  year      = {2024},
}

@inproceedings{levy2023diverse,
  title={Diverse Demonstrations Improve In-context Compositional Generalization},
  author={Levy, Itay and Bogin, Ben and Berant, Jonathan},
  booktitle={Findings of the Association for Computational Linguistics: ACL 2023},
  pages={1401--1422},
  year={2023}
}

@inproceedings{hongjin2022selective,
  title={Selective annotation makes language models better few-shot learners},
  author={Hongjin, SU and Kasai, Jungo and Wu, Chen Henry and Shi, Weijia and Wang, Tianlu and Xin, Jiayi and Zhang, Rui and Ostendorf, Mari and Zettlemoyer, Luke and Smith, Noah A and others},
  booktitle={The Eleventh International Conference on Learning Representations},
  year={2022}
}

@InProceedings{zhang2023ideal,
  author    = {Zhang, Shaokun and Xia, Xiaobo and Wang, Zhaoqing and Chen, Ling-Hao and Liu, Jiale and Wu, Qingyun and Liu, Tongliang},
  booktitle = {The Twelfth International Conference on Learning Representations.},
  title     = {{IDEAL}: Influence-driven selective annotations empower in-context learners in large language models},
  year      = {2023},
}

@inproceedings{gupta2023coverage,
  title={Coverage-based Example Selection for In-Context Learning},
  author={Gupta, Shivanshu and Gardner, Matt and Singh, Sameer},
  booktitle={The 2023 Conference on Empirical Methods in Natural Language Processing},
  year={2023}
}

@InProceedings{NEURIPS2020_1457c0d6,
  author    = {Brown, Tom and Mann, Benjamin and Ryder, Nick and Subbiah, Melanie and Kaplan, Jared D and Dhariwal, Prafulla and Neelakantan, Arvind and Shyam, Pranav and Sastry, Girish and Askell, Amanda and Agarwal, Sandhini and Herbert-Voss, Ariel and Krueger, Gretchen and Henighan, Tom and Child, Rewon and Ramesh, Aditya and Ziegler, Daniel and Wu, Jeffrey and Winter, Clemens and Hesse, Chris and Chen, Mark and Sigler, Eric and Litwin, Mateusz and Gray, Scott and Chess, Benjamin and Clark, Jack and Berner, Christopher and McCandlish, Sam and Radford, Alec and Sutskever, Ilya and Amodei, Dario},
  booktitle = {Advances in Neural Information Processing Systems},
  title     = {Language Models are Few-Shot Learners},
  year      = {2020},
  pages     = {1877--1901},
  volume    = {33},
}

@InProceedings{zhao2021calibrate,
  author       = {Zhao, Zihao and Wallace, Eric and Feng, Shi and Klein, Dan and Singh, Sameer},
  booktitle    = {Thirty-Eighth International Conference on Machine Learning},
  title        = {Calibrate before use: Improving few-shot performance of language models},
  year         = {2021},
  pages        = {12697--12706},
}

@inproceedings{besta2024graph,
  title={Graph of thoughts: Solving elaborate problems with large language models},
  author={Besta, Maciej and Blach, Nils and Kubicek, Ales and Gerstenberger, Robert and Podstawski, Michal and Gianinazzi, Lukas and Gajda, Joanna and Lehmann, Tomasz and Niewiadomski, Hubert and Nyczyk, Piotr and others},
  booktitle={Proceedings of the AAAI Conference on Artificial Intelligence},
  pages={17682--17690},
  year={2024}
}

@Article{friston2008hierarchical,
  author    = {Friston, Karl},
  journal   = {Plos Computational Biology},
  title     = {Hierarchical models in the brain},
  year      = {2008},
  number    = {11},
  pages     = {e1000211},
  volume    = {4},
  publisher = {Public Library of Science San Francisco, USA},
}

@InProceedings{yao2024tree,
  author    = {Yao, Shunyu and Yu, Dian and Zhao, Jeffrey and Shafran, Izhak and Griffiths, Tom and Cao, Yuan and Narasimhan, Karthik},
  booktitle = {Advances in Neural Information Processing Systems},
  title     = {Tree of thoughts: Deliberate problem solving with large language models},
  year      = {2023},
  pages     = {11809--11822},
  volume    = {36},
}

@article{yang2023exploring,
  title={Exploring diverse in-context configurations for image captioning},
  author={Yang, Xu and Wu, Yongliang and Yang, Mingzhuo and Chen, Haokun and Geng, Xin},
  journal={Advances in Neural Information Processing Systems},
  volume={36},
  pages={40924--40943},
  year={2023}
}

@inproceedings{li2024configure,
  title={How to configure good in-context sequence for visual question answering},
  author={Li, Li and Peng, Jiawei and Chen, Huiyi and Gao, Chongyang and Yang, Xu},
  booktitle={Proceedings of the IEEE/CVF Conference on Computer Vision and Pattern Recognition},
  pages={26710--26720},
  year={2024}
}

@article{yang2024lever,
  title={Lever LM: configuring in-context sequence to lever large vision language models},
  author={Yang, Xu and Peng, Yingzhe and Ma, Haoxuan and Xu, Shuo and Zhang, Chi and Han, Yucheng and Zhang, Hanwang},
  journal={Advances in Neural Information Processing Systems},
  volume={37},
  pages={100341--100368},
  year={2024}
}

@article{peng2024live,
  title={Live: Learnable in-context vector for visual question answering},
  author={Peng, Yingzhe and Hu, Xinting and Peng, Jiawei and Geng, Xin and Yang, Xu and others},
  journal={Advances in Neural Information Processing Systems},
  volume={37},
  pages={9773--9800},
  year={2024}
}

@inproceedings{jiang2025mimic,
  title={Mimic In-Context Learning for Multimodal Tasks},
  author={Jiang, Yuchu and Fu, Jiale and Hao, Chenduo and Hu, Xinting and Peng, Yingzhe and Geng, Xin and Yang, Xu},
  booktitle={Proceedings of the Computer Vision and Pattern Recognition Conference},
  pages={29825--29835},
  year={2025}
}

\end{document}